\documentclass[12pt]{article}
\usepackage{fullpage}
\usepackage{times}
\usepackage[normalem]{ulem}
\usepackage{fancyhdr,graphicx,amsmath,amssymb, mathtools, scrextend, titlesec, enumitem}
\usepackage[ruled,vlined]{algorithm2e} 
\usepackage{subcaption}
\usepackage[pagebackref=true,breaklinks=true,colorlinks,bookmarks=false]{hyperref}
\include{pythonlisting}
\include{xspace}

\newcommand{\vthree}{HO-3D\_v3\xspace}
\newcommand{\vtwo}{HO-3D\_v2\xspace}

\newcommand{\cell}[1]{\begin{tabular}{@{}l@{}}#1\end{tabular}}
\newcommand{\nicerresultwidthTwo}{0.4\linewidth}
\newcommand{\nicerrresultTwo}[5]{
\cell{HO-3D\_#5} &
\cell{
\begin{picture}(200,200)
\put(0,0){
\includegraphics[trim=120 50 120 50,clip,width=\nicerresultwidthTwo]{figures/#1.png}}
\put(120,180){\fontsize{15}{6}\selectfont #3}
\end{picture}
} &
\cell{
\begin{picture}(200,200)
\put(0,0){
\includegraphics[trim=120 50 120 50,clip,width=\nicerresultwidthTwo]{figures/#2.png}}
\put(120,180){\fontsize{15}{6}\selectfont #4}
\end{picture}
} \\
}

\title{\textbf{HO-3D\_v3: Improving the Accuracy of Hand-Object Annotations of the HO-3D Dataset}}
\date{}
\author{Shreyas Hampali\textsuperscript{(1)}, Sayan Deb Sarkar\textsuperscript{(1)}, Vincent Lepetit\textsuperscript{(2,1)} \and
\textsuperscript{(1)}{\normalsize Institute for Computer Graphics and Vision, Graz University of Technology, Graz, Austria }\\ 
\textsuperscript{(2)}{\normalsize Universit\'e Paris-Est, \'Ecole des Ponts ParisTech, Paris, France}\\
{\tt\small \{<firstname>.<lastname>\}@icg.tugraz.at} \\
{\tt\small Project page: \href{https://www.tugraz.at/index.php?id=40231}{\color{blue} https://www.tugraz.at/index.php?id=40231}}
}

\begin{document}
\maketitle

\section{What's new in the  `v3' version of HO-3D?}
HO-3D is a dataset providing image sequences of various hand-object interaction scenarios annotated with the 3D pose of the hand and the object and was introduced in \cite{honnotate} as version \vtwo. The annotations were obtained automatically using an optimization method introduced in the original paper.

\vthree provides \textbf{more accurate} annotations for both the hand and object poses thus resulting in \textbf{better estimates of contact regions} between the hand and the object. The new annotations were obtained using an  improved version of the optimization technique of \cite{honnotate}  as detailed in Section~\ref{sec:method} below. Table~\ref{tab:stats} shows the statistics of the \vtwo and \vthree versions of the dataset.

\begin{figure}[h!]
    \centering
    \scalebox{0.75}{
    \begin{tabular}{c|cc}
    \cell{} &
    \cell{Dynamic Grasps (Multi camera)} &
    \cell{Static Grasps (Single camera)}
    \\
    \nicerrresultTwo{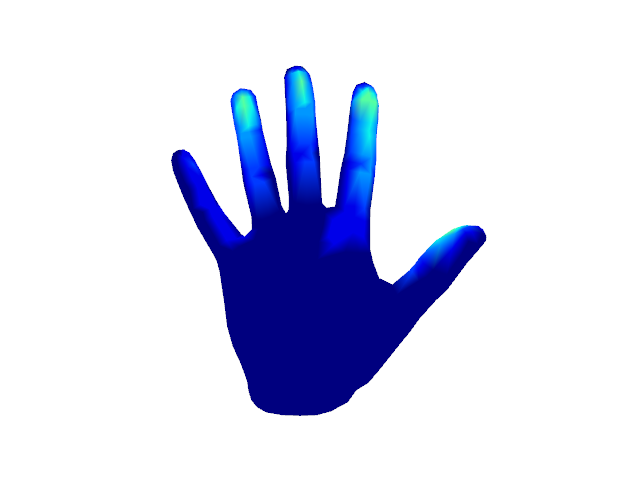}{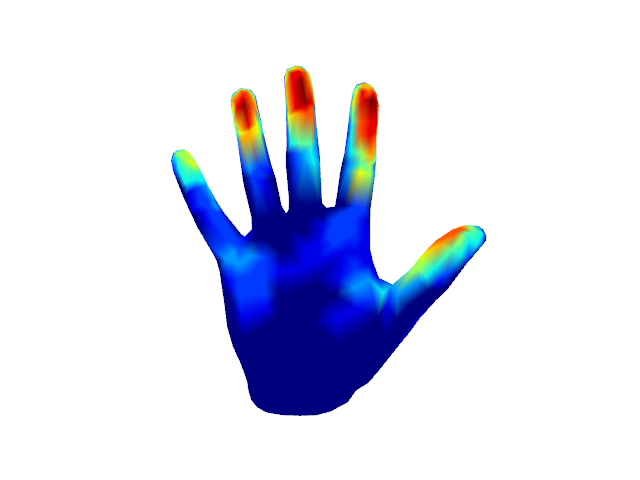}{$1.0\pm 0.9 mm$}{$1.8\pm 0.8 mm$}{v2}
    \hline
    \nicerrresultTwo{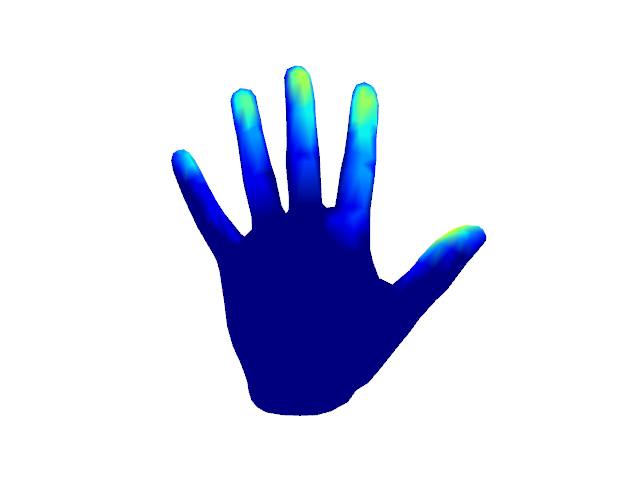}{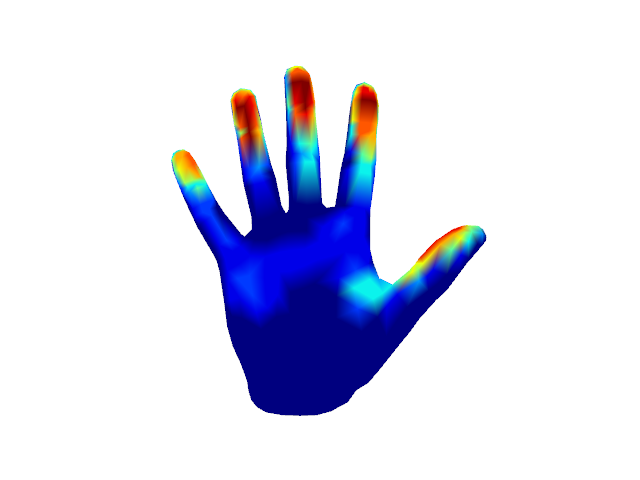}{$1.3\pm 1.0 mm$}{$1.8\pm 0.7 mm$}{v3}
    \end{tabular}
    }
    \caption{Contact maps for the \vtwo and \vthree versions of the datasets for dynamic and static grasps. `Red' regions denote complete contact and `Blue' regions denote no contact. The mean penetration between hand and object are provided in the inset. The `v3' dataset shows more contact regions than the `v2' dataset with significant improvement in the case of dynamic grasps.}
    \label{fig:contact_map_main}
\end{figure}

The improvement in accuracy of \vthree annotations is shown in Table~\ref{tab:accu} based on 50 manually annotated frames. Figure~\ref{fig:contact_map_main} shows the contact regions between the  hand and the objects over the hand surface averaged over the entire dataset together with the mean penetration distance. \vthree annotations clearly results in more contacts compared to \vtwo.

\begin{table}[h]
\centering
\scalebox{1.0}{
\begin{tabular}{c | c | c | c | c | c}
    & Train set & Test Set & &&\\
    & \#images & \#images & \#objects  & \#subjects & Release date\\
    \hline
    \vtwo & 66,034 & 11,524 & 10 & 10 & 9 Jan 2020 \\
    \vthree & 83,325 & 20,137 & 10 & 10 & 1 Jul 2021\\
\end{tabular}
}
\caption{Statistics of the \vtwo and \vthree datasets.}
\label{tab:stats}
\end{table}


\begin{table}[h]
\centering
\begin{tabular}{c | c | c}
            & mean (mm) & std (mm) \\
            \hline
            \vtwo & 12.31 & 3.8  \\
            \vthree & 8.12 & 2.2 \\
\end{tabular}
\caption{Accuracy of our estimations with respect to manual annotations of fingertips using the point cloud built from all camera depth maps. We manually annotated 50 frames with all 5 fingertip locations. The frames were specifically chosen from parts of the sequence where \vtwo provides poor estimates.}
\label{tab:accu}
\end{table}


\section{Algorithmic improvements}
\label{sec:method}

We made the following three changes to our method proposed in \cite{honnotate} in order to  improve the accuracy of the pose estimates.

\subsection{Cross-entropy based Silhouette Discrepancy Term}
We replace the L2 term for the Silhouette error in Eq.~(4) of \cite{honnotate} with the cross-entropy loss. Moreover, instead of obtaining the segmentations for the hand and object from a network trained on synthetic data as in \cite{honnotate}, we use the segmentation prediction by MSeg~\cite{mseg}.

\begin{figure}[h]
\centering
\begin{tabular}{cccc}
  \includegraphics[width=0.22\linewidth]{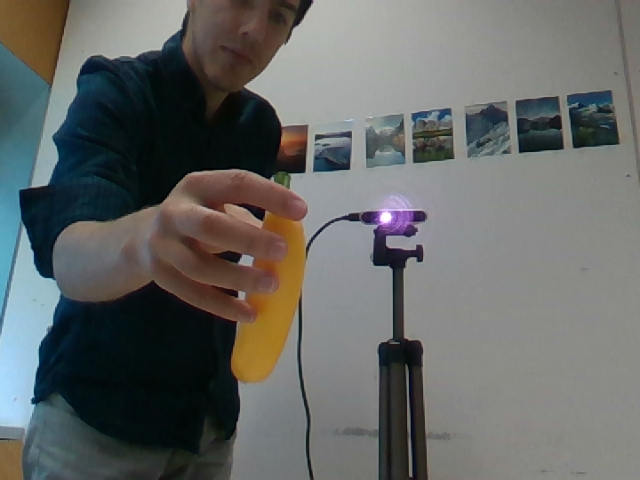} &
  \includegraphics[width=0.22\linewidth]{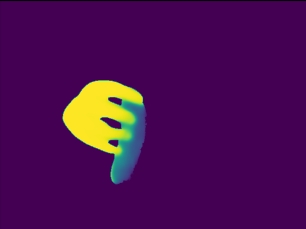} &
  \includegraphics[width=0.22\linewidth]{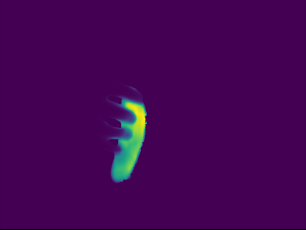} &  
  \includegraphics[width=0.27\linewidth]{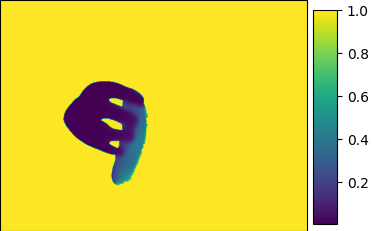} \\
  Color image & Hand & Object & Background \\
  & confidence map & confidence map & confidence map \\
  \end{tabular}
\caption{Confidence maps for the hand, object, and background classes obtained by merging the predictions made by MSeg~\cite{mseg} for 194 categories.}
\label{fig:conf_maps}
\end{figure}

MSeg predicts a segmentation with 194 categories. We remap these categories to 3 classes: \textit{object}, \textit{person} and \textit{background}. Figure~\ref{fig:conf_maps} shows the confidence maps predicted for these 3 categories using the MSeg network on an example.

Denoting the segmentation confidence maps for camera $v$ by $S_v(c)$, where $c \in \textit{\{object, person,}$ $\textit{background\}}$ and the rendered segmentation by $S_v^R(c)$ (\textit{i.e.} a pixel in $S_v^R(c)$ has value 1 if lying on class $c$ and 0 otherwise), the silhouette term is defined as
\begin{equation}
    E_\text{mask} = \sum_c S_v(c)\cdot S_v^R(c) \> .
\end{equation}
%


\subsection{Anatomically Oriented Joint Axes}
\begin{figure}[h]
\centering
\begin{subfigure}{.35\textwidth}
  \centering
  \includegraphics[width=0.8\linewidth]{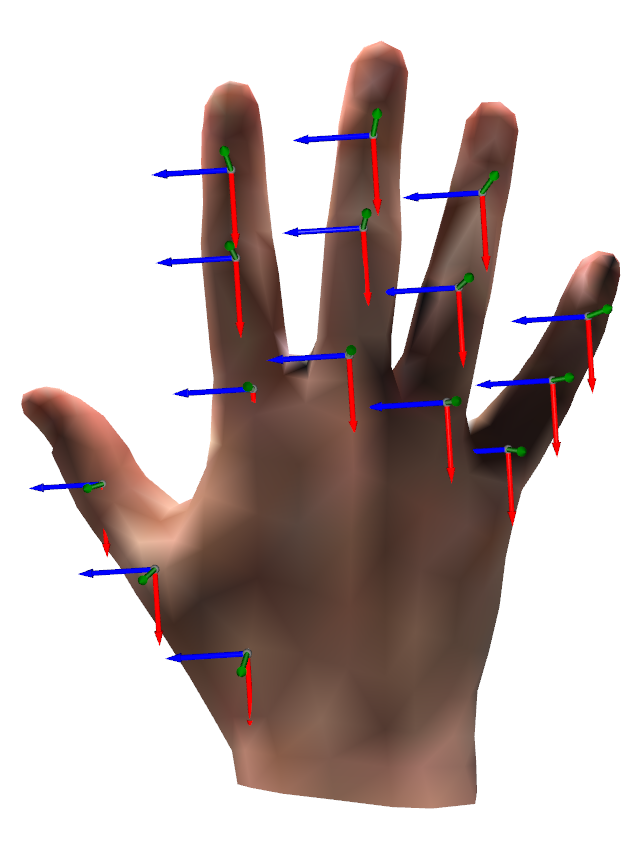}
  \vspace{3px}
  \caption{Joint axes not aligned along anatomical direction in MANO~\cite{mano}}
  \label{fig:jointO}
\end{subfigure}\hspace{5px}
\begin{subfigure}{.35\textwidth}
  \centering
  \includegraphics[width=0.9\linewidth]{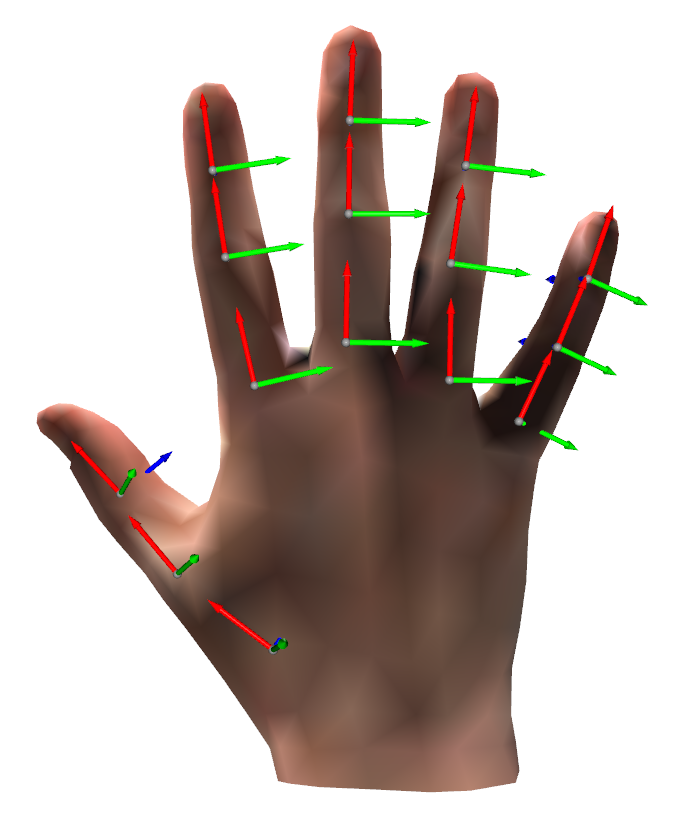}
  \caption{Joint axes manually aligned with joint direction in our method}
  \label{fig:jointN}
\end{subfigure}\\
\begin{subfigure}{.35\textwidth}
  \centering
  \includegraphics[width=1.0\linewidth]{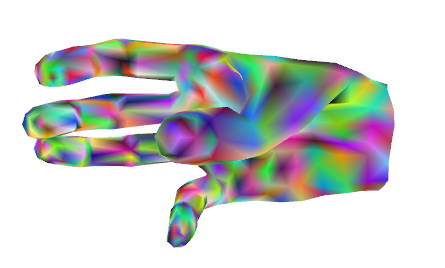}
  \caption{Twisted thumb when using non-anatomically aligned joint axes}
  \label{fig:twistedv2}
\end{subfigure}\hspace{5px}
\begin{subfigure}{.35\textwidth}
  \centering
  \includegraphics[width=1\linewidth]{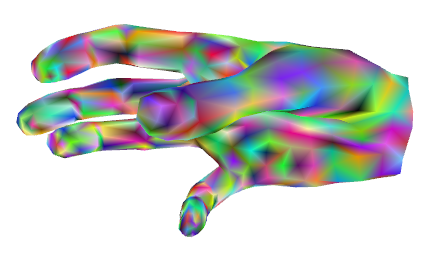}
  \caption{More natural pose when using anatomically aligned joint axes}
  \label{fig:twistedv3}
\end{subfigure}
\caption{We align the joint axes along anatomical direction of the fingers resulting in more natural poses in the \vthree dataset.}
\label{fig:twisted}
\end{figure}

We optimize directly on the joint angles instead of its PCA components. We have found this strategy helpful to 1) reach grasp poses that were not reachable when using PCA components and 2) avoid getting stuck in local minima especially when optimizing on segmentation and depth maps. The joint axes in the MANO~\cite{mano} model are not aligned along the anatomical directions of fingers which makes the task of defining the joint angle limits difficult. Figure~\ref{fig:jointO} shows the joint axes in the MANO model and Figure~\ref{fig:twistedv2} shows the resulting implausible pose due to inaccurate joint angle limits. Figure~\ref{fig:jointN} shows the manually aligned joint axes along the anatomical direction and the resulting pose in the \vthree dataset.


\subsection{Improved Repulsion Term}

\begin{figure}[h]
\centering
\scalebox{0.9}{
 \begin{subfigure}[b]{0.4\textwidth}
        \centering
        \includegraphics[width=0.475\linewidth]{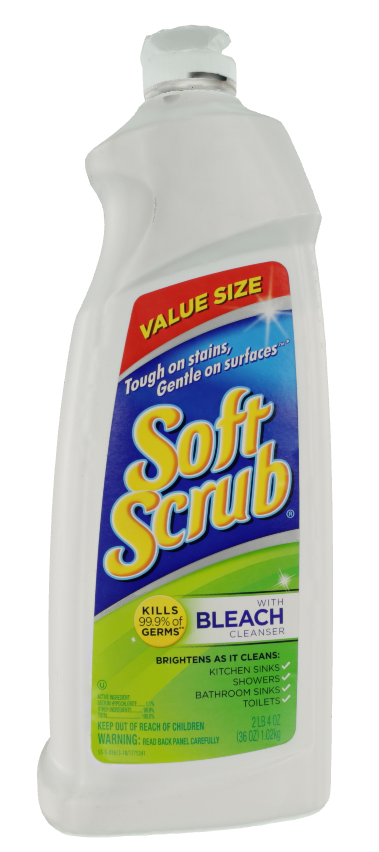}%
        \hfill
        \includegraphics[width=0.42\linewidth]{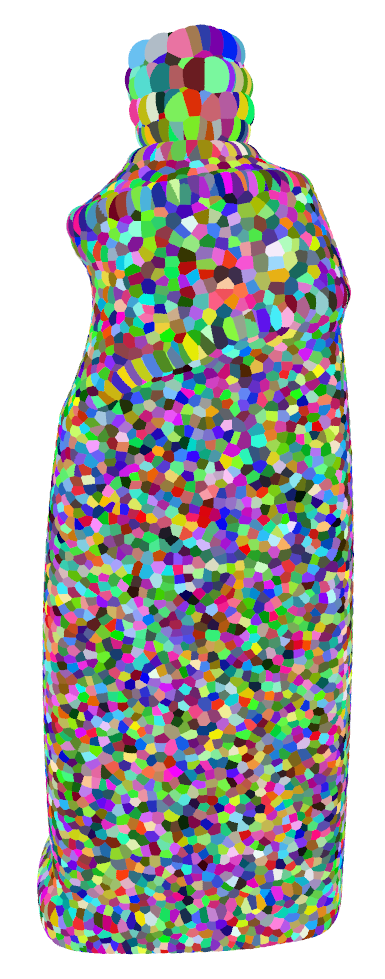}
        \caption{\textit{Bleach Cleanser} object with its inner volume populated with spheres}
    \end{subfigure}%
    \begin{subfigure}[b]{0.5\textwidth}
        \centering
        \includegraphics[width=0.5\linewidth]{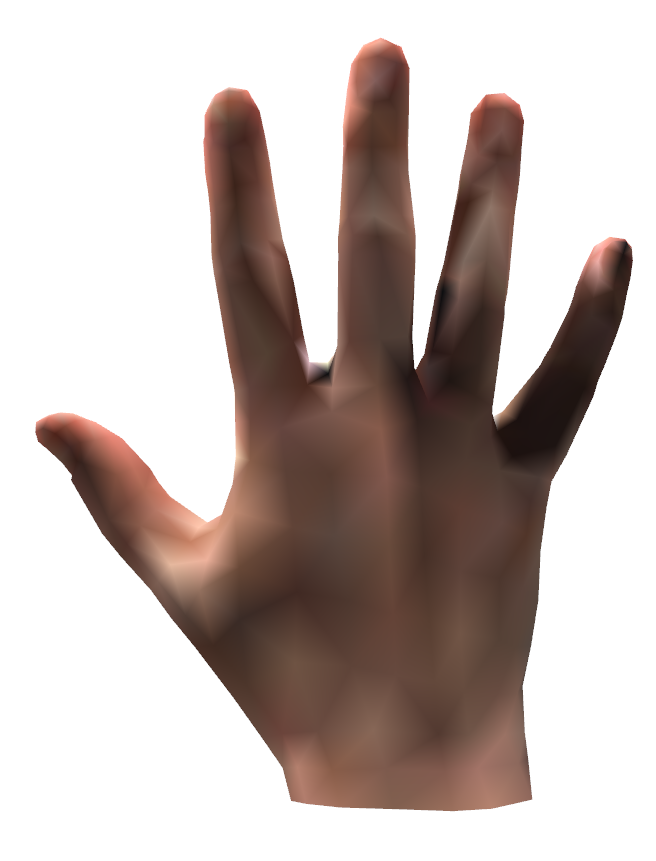}%
        \hfill
        \includegraphics[width=0.5\linewidth]{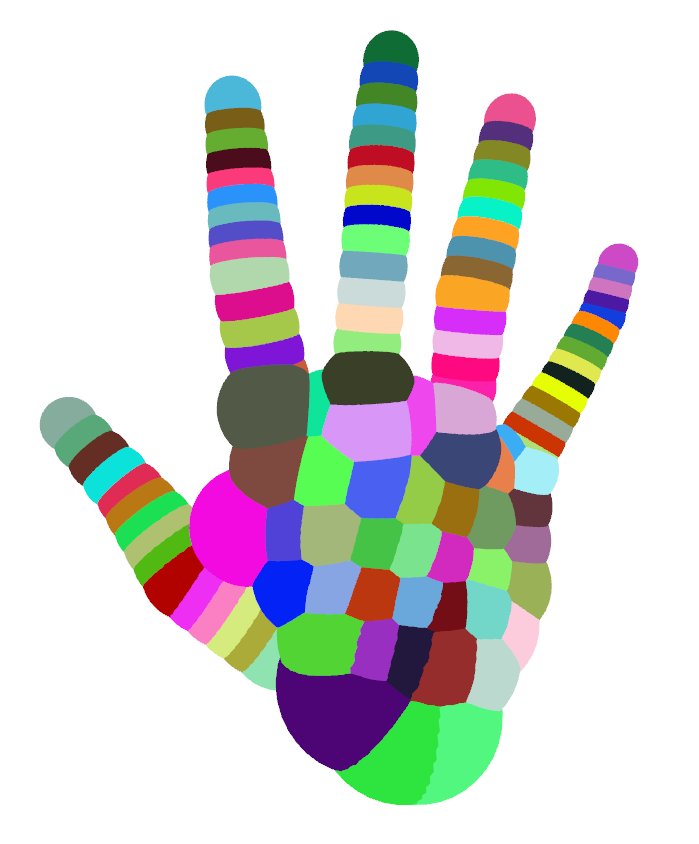}
        \caption{Hand model and its sphere representation\newline}
    \end{subfigure}
}
\caption{Sphere-based object and hand models.}
\label{fig:spheres}
\end{figure}

We replace the physical plausibility term $E_{phy}$ in Eq.~(9) of \cite{honnotate} with a simpler sphere-based repulsion term. More specifically, we obtained a ``sphere-based representation'' for the hand and the object by populating their inner volumes with spheres as shown in Figure~\ref{fig:spheres}.  We define the repulsion term between the $i^\text{th}$ hand sphere and the $j^\text{th}$ object sphere with radii denoted by $r^h_i, r^o_j$ and centers denoted by $c^h_i, c^o_j$, respectively as:
\begin{equation}
    E_\text{phy} = \sum_i \sum_j \max(0, r^h_i+r^o_j - \lvert\lvert c^h_i-c^o_j\lvert\lvert_2 - t) \> ,
\end{equation}
where $t$ is the allowed penetration between the hand and the object. We use $t = 2mm$.


\section{Evaluations}

\begin{figure}[h!]
\centering
\scalebox{0.75}{
 \begin{subfigure}[b]{1.0\textwidth}
        \centering
        \includegraphics[width=0.5\linewidth]{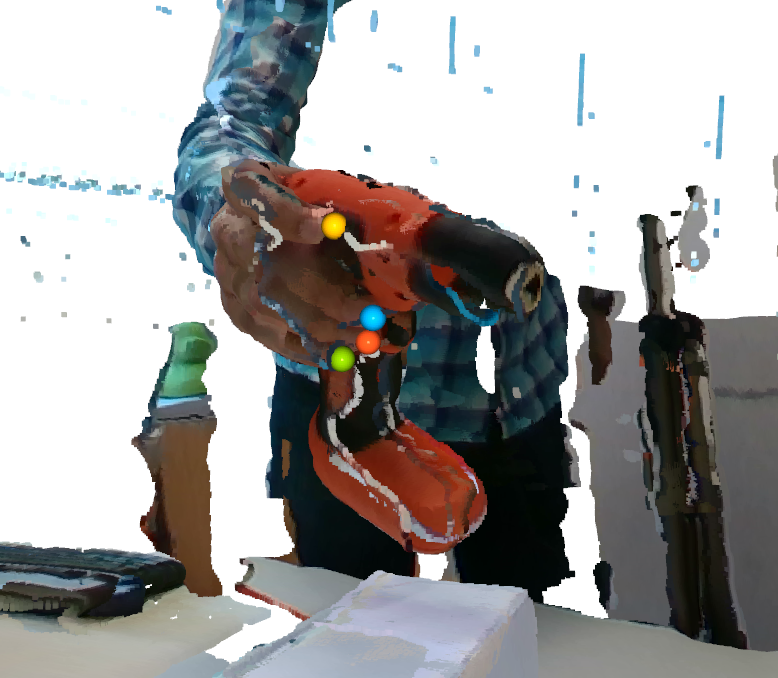}%
        \hfill
        \includegraphics[width=0.5\linewidth]{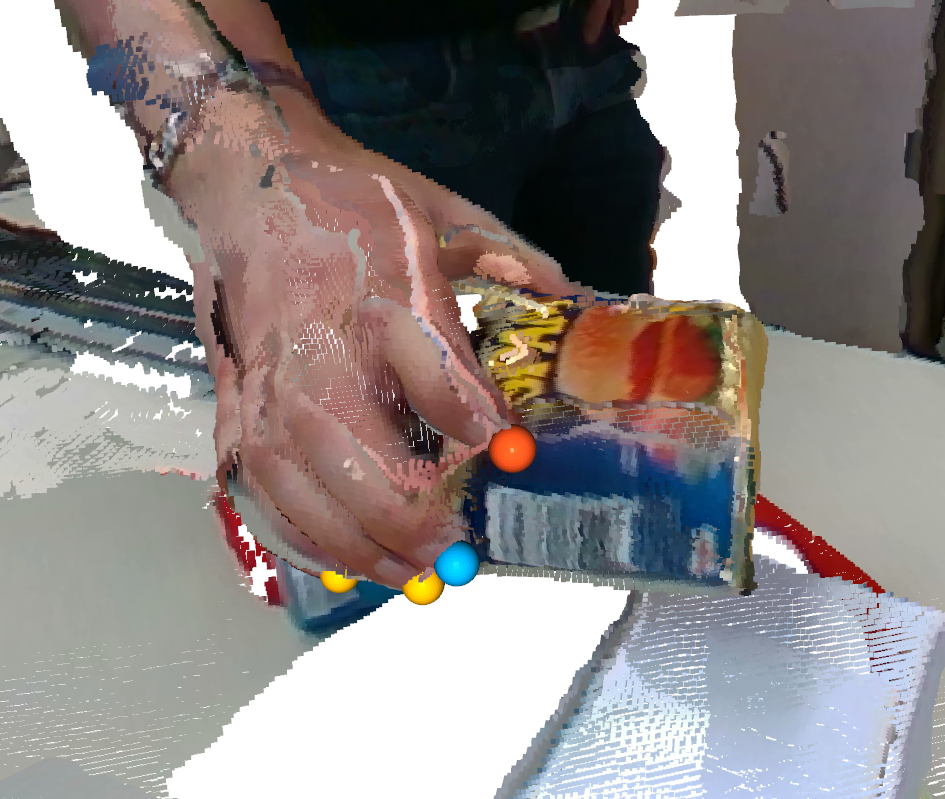}
\end{subfigure}\hspace{5px}
}
\caption{Point cloud from all cameras along with manual annotation of the finger tips.}
\label{fig:pcl}
\end{figure}

We report here the improvement in hand pose accuracy of \vthree  with respect to \vtwo. We manually annotated the point cloud of 50 frames with finger tip locations to measure the accuracy of our method. The point cloud is obtained by combining the point cloud from all cameras in the multi-camera setup as shown in Figure~\ref{fig:pcl}. Further, the 50 frames were specifically chosen from parts of the \vtwo sequences where the hand pose accuracy is poor.  As shown in Table~\ref{tab:accu}, \vthree shows 4mm improvement in accuracy compared to \vtwo. 

Figure~\ref{fig:conf_maps} shows the contact maps for dynamic and static grasp sequences for \vthree and \vtwo datasets along with mean penetration between hand and object. We consider a point on the surface of the hand to be in contact with the object if the distance to the object suface is less than 4mm. We allow this tolerance to account for depth errors and error in camera calibration. The poses in \vthree exhibit more hand-object contact while having similar inter-penetration. The contact maps for each of the sequences in the dataset are shown in Figures~\ref{fig:cont_maps1}, \ref{fig:cont_maps2} and \ref{fig:cont_maps3}.

\bibliographystyle{unsrt}
\bibliography{sample}


\newcommand{\intercompwidth}{0.3\linewidth}
\newcommand{\contResult}[1]{
\begin{minipage}{0.1\linewidth}
\centering
#1
\end{minipage}%
\begin{minipage}{\intercompwidth}
\centering
\subfloat{\includegraphics[width=1\linewidth]{figures/contact_map_hand/#1/v2/contact_map_hand_right_0.003000.jpg}}
\end{minipage}%
\begin{minipage}{\intercompwidth}
\centering
\subfloat{\includegraphics[width=1\linewidth]{figures/contact_map_hand/#1/v3/contact_map_hand_right_0.003000.jpg}}
\end{minipage}%
\par\medskip
}

\newcommand{\nicerresultwidth}{0.2\linewidth}
\newcommand{\nicerrresultFour}[9]{
\cell{HO-3D\_#9} &
\cell{
\begin{picture}(100,100)
\put(0,0){
\includegraphics[trim=120 50 120 50,clip,width=\nicerresultwidth]{figures/contact_map_hand/#1/#9/contact_map_hand_right_0.003000.jpg}}
\put(60,90){\fontsize{10}{6}\selectfont #5}
\end{picture}
} &
\cell{
\begin{picture}(100,100)
\put(0,0){
\includegraphics[trim=120 50 120 50,clip,width=\nicerresultwidth]{figures/contact_map_hand/#2/#9/contact_map_hand_right_0.003000.jpg}}
\put(60,90){\fontsize{10}{6}\selectfont #6}
\end{picture}
} &
\cell{
\begin{picture}(100,100)
\put(0,0){
\includegraphics[trim=120 50 120 50,clip,width=\nicerresultwidth]{figures/contact_map_hand/#3/#9/contact_map_hand_right_0.003000.jpg}}
\put(60,90){\fontsize{10}{6}\selectfont #7}
\end{picture}
} &
\cell{
\begin{picture}(100,100)
\put(0,0){
\includegraphics[trim=120 50 120 50,clip,width=\nicerresultwidth]{figures/contact_map_hand/#4/#9/contact_map_hand_right_0.003000.jpg}}
\put(60,90){\fontsize{10}{6}\selectfont #8}
\end{picture}
} \\
}

\begin{figure*}
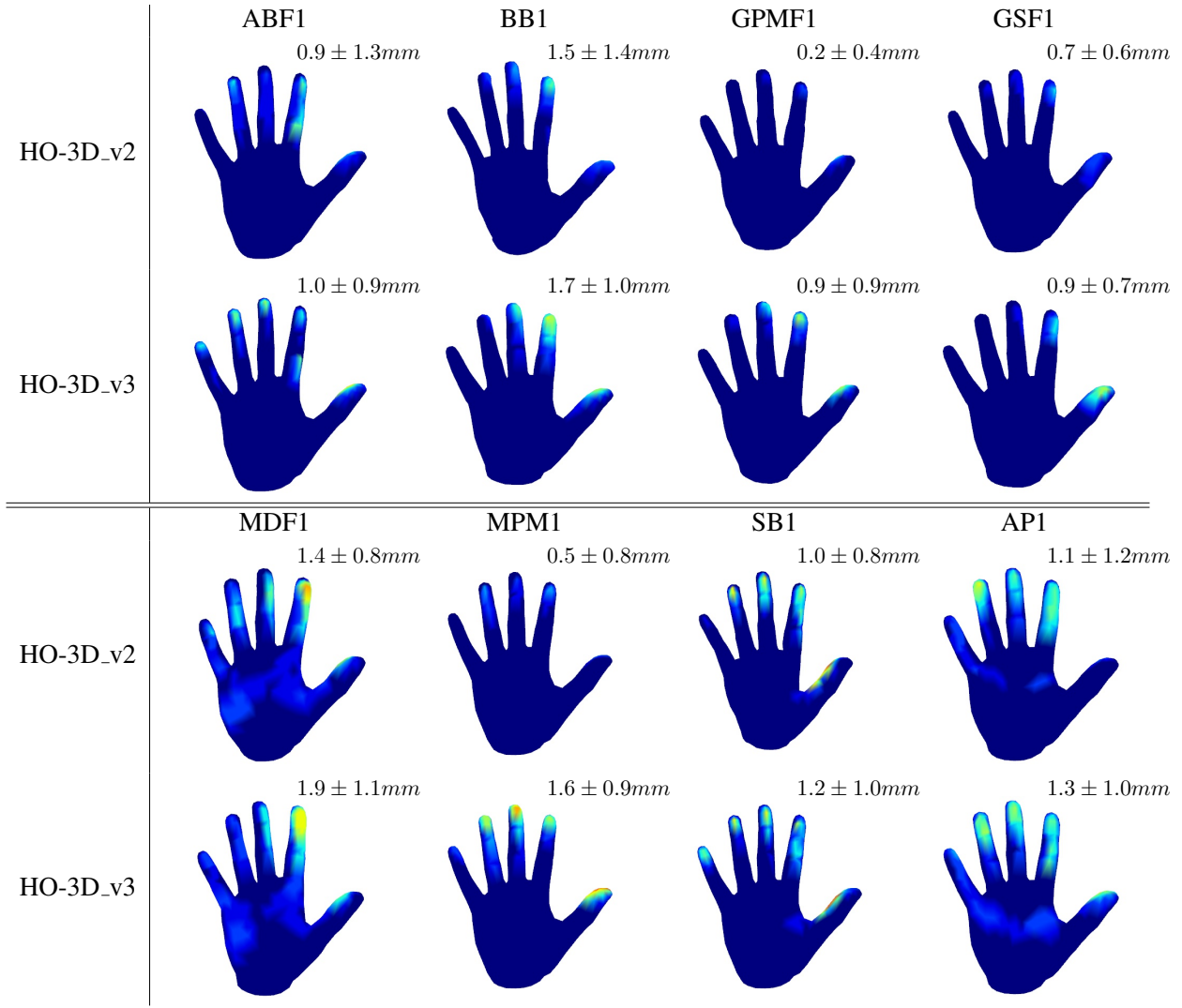

    \centering
    \scalebox{0.9}{
    \begin{tabular}{c|cccc}

        \cell{} &
        \cell{ABF1} &
        \cell{BB1} &
        \cell{GPMF1} &
        \cell{GSF1}
        \\
        \nicerrresultFour{ABF1}{BB1}{GPMF1}{GSF1}{$0.9\pm 1.3 mm$}{$1.5\pm 1.4 mm$}{$0.2\pm 0.4 mm$}{$0.7\pm 0.6 mm$}{v2}
        \nicerrresultFour{ABF1}{BB1}{GPMF1}{GSF1}{$1.0\pm 0.9 mm$}{$1.7\pm 1.0 mm$}{$0.9\pm 0.9 mm$}{$0.9\pm 0.7 mm$}{v3}
        \hline
        \hline
        \cell{} &
        \cell{MDF1} &
        \cell{MPM1} &
        \cell{SB1} &
        \cell{AP1}
        \\
        \nicerrresultFour{MDF1}{MPM1}{SB1}{AP1}{$1.4\pm 0.8 mm$}{$0.5 \pm 0.8 mm$}{$1.0\pm 0.8 mm$}{$1.1\pm 1.2 mm$}{v2}
        \nicerrresultFour{MDF1}{MPM1}{SB1}{AP1}{$1.9\pm 1.1 mm$}{$1.6 \pm 0.9 mm$}{$1.2\pm 1.0 mm$}{$1.3\pm 1.0 mm$}{v3}
       
    \end{tabular}
    }
    \caption{Comparison of contact maps between the `v2' and `v3' versions of the HO-3D dataset for all the dynamic grasp sequences. The mean penetration for each sequence is provided in the inset. We see an overall improvement in grasp quality.}
    \label{fig:cont_maps1}
\end{figure*}

\begin{figure*}
    \centering
    \scalebox{0.8}{
    \begin{tabular}{c|cccc}

        \cell{} &
        \cell{SM1} &
        \cell{SM2} &
        \cell{SM3} &
        \cell{SM4}
        \\
        \nicerrresultFour{SM1}{SM2}{SM3}{SM4}{$2.2 mm$}{$3.0 mm$}{$2.4 mm$}{$1.3 mm$}{v2}
        \nicerrresultFour{SM1}{SM2}{SM3}{SM4}{$1.7 mm$}{$1.8 mm$}{$1.5 mm$}{$1.3 mm$}{v3}
        \hline
        \hline
        \cell{} &
        \cell{SM5} &
        \cell{MC1} &
        \cell{MC2} &
        \cell{MC4}
        \\
        \nicerrresultFour{SM5}{MC1}{MC2}{MC4}{$2.3 mm$}{$0.8 mm$}{$1.9 mm$}{$1.6 mm$}{v2}
        \nicerrresultFour{SM5}{MC1}{MC2}{MC4}{$1.2 mm$}{$2.0 mm$}{$2.7 mm$}{$2.1 mm$}{v3}
        \hline
        \hline
        \cell{} &
        \cell{MC5} &
        \cell{MC6} &
        \cell{SS1} &
        \cell{SS2}
        \\
        \nicerrresultFour{MC5}{MC6}{SS1}{SS2}{$2.8 mm$}{$1.1 mm$}{$1.4 mm$}{$2.1 mm$}{v2}
        \nicerrresultFour{MC5}{MC6}{SS1}{SS2}{$3.2 mm$}{$1.9 mm$}{$1.4 mm$}{$2.1 mm$}{v3}
       
    \end{tabular}
    }
    \caption{Comparison of contact maps between the `v2' and `v3' versions of the HO-3D dataset for the static grasp sequences. The mean penetration for each sequence is provided in the inset. The grasp quality between the two versions are similar.}
    \label{fig:cont_maps2}
\end{figure*}

\begin{figure*}
    \centering
    \scalebox{0.9}{
    \begin{tabular}{c|cccc}
        \cell{} &
        \cell{SS3} &
        \cell{SiS1} &
        \cell{ND2} &
        \cell{SMu1}
        \\
        \nicerrresultFour{SS3}{SiS1}{ND2}{SMu1}{$2.9 mm$}{$0.9 mm$}{$2.5 mm$}{$0.4 mm$}{v2}
        \nicerrresultFour{SS3}{SiS1}{ND2}{SMu1}{$2.3 mm$}{$1.0 mm$}{$2.5 mm$}{$0.3 mm$}{v3}
        \hline
        \hline
        \cell{} &
        \cell{SMu4}
        \\
        \cell{\vtwo} &
        \cell{
        \begin{picture}(100,100)
        \put(0,0){
        \includegraphics[trim=120 50 120 50,clip,width=\nicerresultwidth]{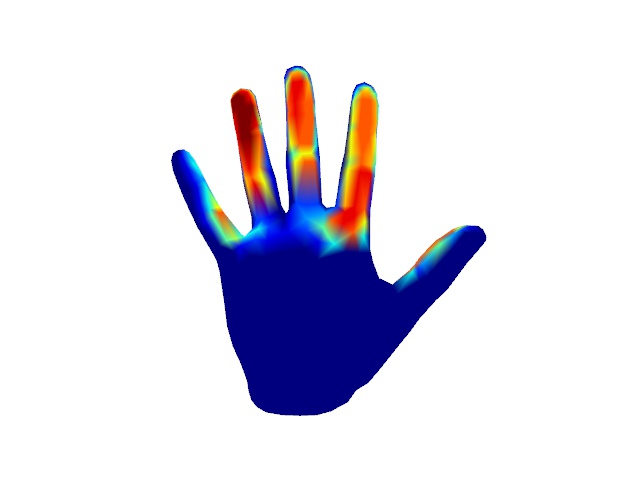}}
        \put(60,90){\fontsize{10}{6}\selectfont $1.4 mm$}
        \end{picture}
        }\\
        \cell{\vthree} &
        \cell{
        \begin{picture}(100,100)
        \put(0,0){
        \includegraphics[trim=120 50 120 50,clip,width=\nicerresultwidth]{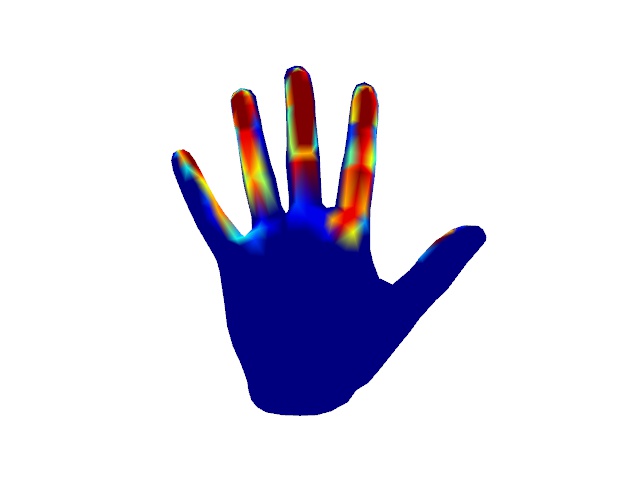}}
        \put(60,90){\fontsize{10}{6}\selectfont $0.4 mm$}
        \end{picture}
        }
       
    \end{tabular}
    }
    \caption{Comparison of contact maps between the `v2' and `v3' versions of the HO-3D dataset for the static grasp sequences. The mean penetration for each sequence is provided in the inset. The grasp quality between the two versions are similar.}
    \label{fig:cont_maps3}
\end{figure*}

\end{document}